\pdfoutput=1

\documentclass[11pt]{article}

\usepackage{EMNLP2022}

\usepackage{times}
\usepackage{latexsym}

\usepackage[T1]{fontenc}

\usepackage[utf8]{inputenc}

\usepackage{microtype}

\usepackage{inconsolata}

\usepackage{microtype}
\usepackage{hyperref}
\usepackage{url}

\usepackage{booktabs}       
\usepackage{amsfonts}       
\usepackage{xcolor}         

\usepackage{graphicx}
\usepackage{subfigure}
\usepackage{amssymb}
\usepackage{amsmath}
\usepackage{amsthm}

\usepackage{multirow}
\usepackage{algorithm}
\usepackage{algorithmic}

\makeatletter
\newsavebox{\@brx}
\newcommand{\llangle}[1][]{\savebox{\@brx}{\(\m@th{#1\langle}\)}%
  \mathopen{\copy\@brx\kern-0.5\wd\@brx\usebox{\@brx}}}
\newcommand{\rrangle}[1][]{\savebox{\@brx}{\(\m@th{#1\rangle}\)}%
  \mathclose{\copy\@brx\kern-0.5\wd\@brx\usebox{\@brx}}}

\title{Complex Hyperbolic Knowledge Graph Embeddings \\ with Fast Fourier Transform}

\author{Huiru Xiao \and Xin Liu \and Yangqiu Song \\
  Hong Kong University of Science and Technology, Hong Kong SAR \\
  \texttt{\{hxiaoaf,xliucr,yqsong\}@cse.ust.hk} \\
  \AND
  Ginny Y. Wong \and Simon See \\
  NVIDIA AI Technology Center (NVATIC), NVIDIA, Santa Clara, USA \\
  \texttt{\{gwong,ssee\}@nvidia.com} \\} 

\begin{document}
\maketitle
\begin{abstract}
The choice of geometric space for knowledge graph (KG) embeddings can have significant effects on the performance of KG completion tasks. The hyperbolic geometry has been shown to capture the hierarchical patterns due to its tree-like metrics, which addressed the limitations of the Euclidean embedding models. Recent explorations of the complex hyperbolic geometry further improved the hyperbolic embeddings for capturing a variety of hierarchical structures. However, the performance of the hyperbolic KG embedding models for non-transitive relations is still unpromising, while the complex hyperbolic embeddings do not deal with multi-relations. This paper aims to utilize the representation capacity of the complex hyperbolic geometry in multi-relational KG embeddings. To apply the geometric transformations which account for different relations and the attention mechanism in the complex hyperbolic space, we propose to use the fast Fourier transform (FFT) as the conversion between the real and complex hyperbolic space. Constructing the attention-based transformations in the complex space is very challenging, while the proposed Fourier transform-based complex hyperbolic approaches provide a simple and effective solution. Experimental results show that our methods outperform the baselines, including the Euclidean and the real hyperbolic embedding models.
\end{abstract}

\section{Introduction}

\begin{figure}[t]
\begin{center}
{\includegraphics[width=1.0\columnwidth]{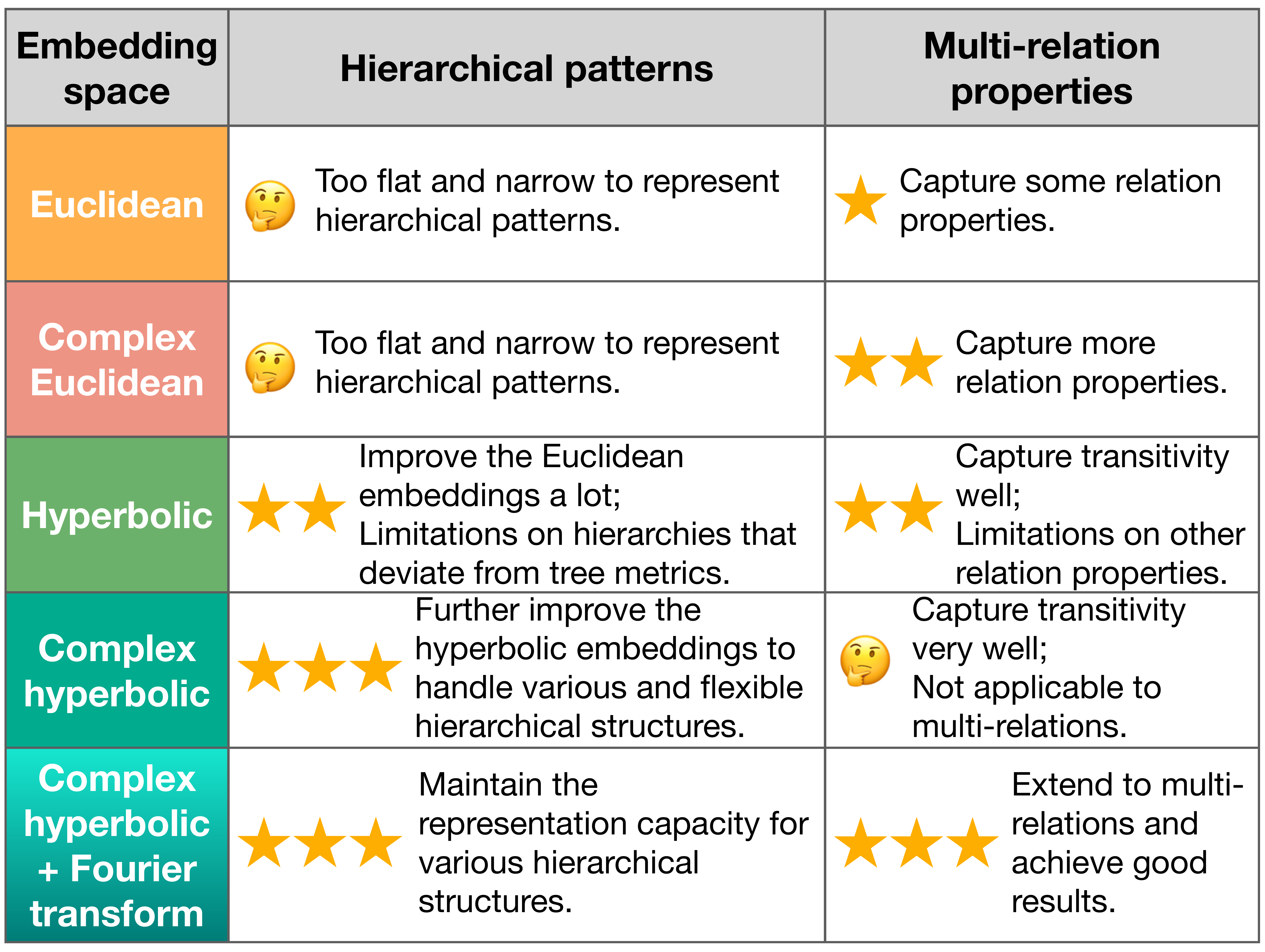}}
\caption{The summary of embedding spaces for hierarchical patterns and multi-relation properties.}
\label{fig:embedding space comparison}
\end{center}
\end{figure}

Knowledge graph (KG) representation learning is important to the KG inference as well as the downstream tasks~\cite{DBLP:journals/pieee/Nickel0TG16}. It has been noticed that the embedding space has significant effects on the {performance} of KG completion tasks. Previous works have proposed the KG embedding models in Euclidean space~\cite{DBLP:conf/nips/BordesUGWY13,DBLP:conf/icml/NickelTK11,DBLP:journals/corr/YangYHGD14a}, complex Euclidean space~\cite{DBLP:conf/icml/TrouillonWRGB16,DBLP:conf/iclr/SunDNT19}, hyperbolic space~\cite{DBLP:conf/nips/BalazevicAH19,DBLP:conf/acl/ChamiWJSRR20}.\footnote{To avoid wordiness, in this paper, we use \textit{hyperbolic space} to refer to \textit{real hyperbolic space}, \textit{hyperbolic geometry} to refer to \textit{real hyperbolic geometry}, and \textit{hyperbolic embeddings} to refer to \textit{real hyperbolic embeddings}.} These models learn the embeddings of the KG entities in the selected geometric spaces and parameterize the relation representations as the geometric transformations, such as translation, rotation, matrix multiplication, etc. 

The Euclidean and complex Euclidean embedding models can capture relation properties including symmetry/anti-symmetry, inversion, and composition, but they cannot handle the transitive relations such as hypernymy. Generally, the transitive relation forms a tree-like structure, for which hyperbolic geometry has a more powerful representation capacity than Euclidean geometry because the hyperbolic space can be regarded as a continuous approximation to trees~\cite{krioukov2010hyperbolic}. 

However, most real-world graphs with transitivity do not necessarily form exact tree structures since the transitive relations can lead to a globally hierarchical structure with varying local structures, such as multitree structures~\cite{DBLP:journals/jct/GriggsLL12} and taxonomies~\cite{DBLP:conf/www/SuchanekKW07}. Thus, the hyperbolic geometry which resembles tree metrics still has limitations on capturing various and flexible hierarchical structures. To tackle the limitation of hyperbolic embeddings, a recent work~\cite{xiao2021unit} proposed to explore the complex hyperbolic geometry to learn the embeddings of hierarchical graphs. Due to the variable negative curvature~\cite{goldman1999complex}, the complex hyperbolic space is more flexible in handling varying structures while the tree-like properties are still retained. 
Despite the remarkable improvements in single transitive relation inference, the complex hyperbolic geometry has not been utilized for multi-relational embeddings.

In this paper, we are motivated to make use of the complex hyperbolic geometry's representation superiority in KGs. There are two main challenges in extending the complex hyperbolic embeddings to multiple relations. First, the geometric transformations in complex hyperbolic geometry are complicated and challenging to optimize due to the numerical instabilities, making it difficult to apply the complex geometric transformations for different relations. Second, it is hard to build the neural network unit or layer in the complex domain. Missing the complex attention mechanism would restrict the parameterization capability and make the complex domain-based model difficult to generalize to further downstream tasks. 

To address the above problems, we propose a complex hyperbolic KG embedding approach with the fast Fourier transform. Our approach can utilize the representation capacity of the complex hyperbolic geometry as well as the well-developed attention-based geometric transformations as relation parameterization, while we borrow the fast Fourier transform (FFT) and inverse fast Fourier transform (IFFT) to provide the conversion between the real and complex hyperbolic space.
We regard the complex hyperbolic embeddings in the unit ball model (a projective geometry-based model to identify the complex hyperbolic space)~\cite{goldman1999complex} and the hyperbolic embeddings in the Poincar\'e ball model (a model of the real hyperbolic space)~\cite{cannon1997hyperbolic} as frequency domain and {spatial} domain respectively. Then FFT and IFFT enable us to convert the embeddings between the two geometric spaces, accomplishing the leverage of real hyperbolic transformations to the complex hyperbolic model.
The framework is simple and effective {in learning} the complex hyperbolic KG representations.

Figure \ref{fig:embedding space comparison} summarizes the comparison among embedding spaces for hierarchical patterns and multi-relation properties. In experiments, we evaluate our approach on the KG link prediction task with two popular benchmarks---WN18RR~\cite{DBLP:conf/nips/BordesUGWY13} and FB15k-237~\cite{toutanova2015observed}. Empirical results show that our Fourier transform-based complex hyperbolic KG embedding approach outperforms the baseline models in other geometric spaces. 

The code and data of our work are available at \url{https://github.com/HKUST-KnowComp/ComplexHyperbolicKGE}.

\section{Related Work}
\paragraph{Euclidean KG embeddings.} The traditional KG embedding models first started with the Euclidean geometry because of its convenient vectorial structure and closed-form computations such as distance formula and inner-product. After the occurrence of the translation-based models~\cite{DBLP:conf/nips/BordesUGWY13} and bilinear models~\cite{DBLP:conf/icml/NickelTK11,DBLP:journals/corr/YangYHGD14a}, several extensions~\cite{DBLP:conf/aaai/WangZFC14,DBLP:conf/aaai/LinLSLZ15,DBLP:conf/acl/JiHXL015} have been made to further develop the Euclidean methods.

\paragraph{Complex Euclidean KG embeddings.} The follow-up works~\cite{DBLP:conf/icml/TrouillonWRGB16,DBLP:conf/acl/HayashiS17,DBLP:conf/iclr/SunDNT19} extended the traditional Euclidean models to complex hyperbolic geometry. Specifically, ComplEx~\cite{DBLP:conf/icml/TrouillonWRGB16} found that the Hermitian dot product can effectively capture anti-symmetric relations while retaining the efficiency benefits of the dot product. RotatE~\cite{DBLP:conf/iclr/SunDNT19} defined each relation as a rotation in the complex vector space to infer various relation patterns (symmetry/anti-symmetry, inversion, composition). The effectiveness of these models revealed the potential of the complex geometry.

\paragraph{Hyperbolic embeddings.} In recent years, the hyperbolic space attracted much attention for representation learning since it can naturally characterize tree structures. The hyperbolic embedding methods have developed from the single transitive relation graphs~\cite{DBLP:conf/nips/NickelK17,DBLP:conf/icml/NickelK18,DBLP:conf/nips/SonthaliaG20} to multi-relational KGs~\cite{DBLP:conf/nips/BalazevicAH19,DBLP:conf/acl/ChamiWJSRR20}. MurP~\cite{DBLP:conf/nips/BalazevicAH19} embedded the hierarchical multi-relational data in the Poincaré ball model and learned relation-specific parameters by Möbius operations. The state-of-the-art hyperbolic KG embedding models are a series of hyperbolic transformation-based models RefH, RotH, and AttH~\cite{DBLP:conf/acl/ChamiWJSRR20}, which utilize the geometric tree-like property to capture the hierarchical structure naturally while using different geometric transformations as well as attention mechanism to parameterize other relation properties.

\paragraph{Lightweight Euclidean-based models.} Based on the hyperbolic embedding model RotH~\cite{DBLP:conf/acl/ChamiWJSRR20}, \citet{DBLP:conf/emnlp/WangLLS21} developed two lightweight Euclidean-based models RotL and Rot2L, which simplified the hyperbolic operations while keeping the flexible normalization effect. 

\paragraph{Complex hyperbolic embeddings.} Since many real-world hierarchically structured data such as taxonomies~\cite{DBLP:journals/cacm/Miller95,DBLP:conf/www/SuchanekKW07} and multitree networks~\cite{DBLP:journals/jct/GriggsLL12} have varying local structures, they do not ubiquitously match the hyperbolic geometry. Therefore, \citet{xiao2021unit} explored the complex hyperbolic space to embed a variety of hierarchical structures. The complex hyperbolic embedding approach improved over the hyperbolic embedding models, but it only focused on the representation of single-relational graphs instead of multi-relational KGs.

\paragraph{Fourier Transform.} Fourier transform~\cite{heideman1984gauss} converts a finite-sequence signal from its temporal or {spatial} domain to the frequency domain. 
FFT~\cite{cooley1969finite} is a practical algorithm that computes the discrete Fourier transform (DFT) of a sequence. FFT and inverse FFT are widely used for many applications~\cite{DBLP:journals/cse/Rockmore00,burgess2014history} for their usefulness in signal processing as well as computation efficiency. They are also used to efficiently perform operations such as convolutions~\cite{smith1997scientist,DBLP:conf/iclr/KipfW17} and cross-correlations~\cite{bracewell1986fourier,DBLP:conf/aaai/WangZXY18}. \citet{DBLP:conf/emnlp/HayashiS19} also introduced the Fourier transform in KGE, where the main idea was to use the block circulant matrices to parameterize relations. While in our work, the Fourier transform is used to transform the entity embeddings between different geometric spaces.

\section{Preliminaries}
\subsection{Hyperbolic Geometry}
\label{sec:hyperbolic geometry}
The hyperbolic space is a homogeneous space with constant negative curvature~\cite{cannon1997hyperbolic}.
In the hyperbolic space, the volume of a ball grows exponentially with its radius.
Contrastively, in the Euclidean space, the curvature is constantly $0$, and the volume of a ball grows polynomially with its radius.
The exponential volume growth rate enables the hyperbolic space to have powerful representation capability for tree structures since the number of nodes grows exponentially with the depth in a tree, while the Euclidean space is too flat and narrow to embed trees.


\paragraph{The Poincar\'e Ball Model.}
To describe the hyperbolic space in mathematical language, there are several models, among which the Poincar\'e ball model is popular for graph representation~\cite{DBLP:conf/nips/NickelK17,DBLP:conf/acl/ChamiWJSRR20} due to the relatively convenient computations.

Denote the Poincar\'e ball model with constant negative curvature $-c$ as $\mathcal{P}_\mathbb{R}^N=\{\mathbf{x}\in\mathbb{R}^N:\|\mathbf{x}\|^2<\frac{1}{c}\}$, which represents the open $N$-dimension ball in the ambient Euclidean space ($\|\cdot\|$ is the Euclidean $L_2$ norm). 
By the framework of gyrovector space~\cite{ungar2008analytic}, the hyperbolic space can be formalized as an approximated vectorial structure, where the M\"{o}bius addition~\cite{DBLP:conf/nips/GaneaBH18} is used as the vector addition in $\mathcal{P}_\mathbb{R}^N$: \\
\scalebox{0.90}{\parbox{1.111\linewidth}{
\begin{equation}
\label{eq:Mobius addition}
    \mathbf{x}\oplus_c\mathbf{y}=\frac{(1+2c\mathbf{x}\mathbf{y}+c\|\mathbf{y}\|^2)\mathbf{x}+(1-c\|\mathbf{x}\|^2)\mathbf{y}}{1+2c\mathbf{x}\mathbf{y}+c^2\|\mathbf{x}\|^2\|\mathbf{y}\|^2}.
\end{equation}
}}
Then the distance function in $\mathcal{P}_\mathbb{R}^N$ is given by \\
\scalebox{0.90}{\parbox{1.111\linewidth}{
\begin{equation}
    d_\mathcal{P}(\mathbf{x},\mathbf{y})=\frac{2}{\sqrt{c}}\text{artanh}(\sqrt{c}\|-\mathbf{x}\oplus_c\mathbf{y}\|).
\end{equation}
}}

The practical computations in the hyperbolic space are often implemented using the tangent space. For $\mathbf{x}\in\mathcal{P}_\mathbb{R}^N$, the associated tangent space $\mathcal{T}_\mathbf{x}\mathcal{P}_\mathbb{R}^N$ is {an} $N$-dimension Euclidean space containing all tangent vectors passing through $\mathbf{x}$~\cite{DBLP:books/daglib/0090942}. The manifold of the Poincar\'e ball model and the tangent space have closed-form maps to each other, which are defined as the exponential map $\exp_\mathbf{0}^c(\mathbf{v}):\mathcal{T}_\mathbf{0}\mathcal{P}_\mathbb{R}^N\mapsto\mathcal{P}_\mathbb{R}^N$ and the logarithmic map $\log_\mathbf{0}^c(\mathbf{y}):\mathcal{P}_\mathbb{R}^N\mapsto\mathcal{T}_\mathbf{0}\mathcal{P}_\mathbb{R}^N$: \\
\scalebox{0.90}{\parbox{1.111\linewidth}{
\begin{align}
\label{eq:exp map}
    &\exp_\mathbf{0}^c(\mathbf{v})=\tanh(\sqrt{c}\|\mathbf{v}\|)\frac{\mathbf{v}}{\sqrt{c}\|\mathbf{v}\|}, \\
    \label{eq:log map}
    &\log_\mathbf{0}^c(\mathbf{y})=\text{artanh}(\sqrt{c}\|\mathbf{y}\|)\frac{\mathbf{y}}{\sqrt{c}\|\mathbf{y}\|}.
\end{align}
}}

\subsection{Complex Hyperbolic Geometry}
The complex hyperbolic space is a homogeneous space of variable negative curvature~\cite{goldman1999complex}. The complex hyperbolic space also maintains the tree-like exponential volume growth property. From the properties of the complex hyperbolic geometry, we see that the complex hyperbolic space can naturally handle data with diverse local structures because of the non-constant curvature while preserving the tree-like properties to better capture the transitivity~\cite{xiao2021unit}.

\paragraph{The Unit Ball Model.}
The complex hyperbolic space's ambient Hermitian vector space $\mathbb{C}^{n,1}$ is the complex Euclidean space $\mathbb{C}^{n+1}$ endowed with some Hermitian form $\llangle\mathbf{z},\mathbf{w}\rrangle$, where $\mathbf{z},\mathbf{w}\in\mathbb{C}^{n+1}$. Different choices of the Hermitian form $\llangle\mathbf{z},\mathbf{w}\rrangle$ correspond to different models of complex hyperbolic geometry, such as the unit ball model and the Siegel domain model~\cite{parker2003notes}. Here we choose the standard Hermitian form which derives the unit ball model: \\
\scalebox{0.90}{\parbox{1.111\linewidth}{
\begin{equation}
\label{eq:Hermitian form}
    \llangle\mathbf{z},\mathbf{w}\rrangle=z_1\overline{w_1}+\cdots+z_n\overline{w_n}-z_{n+1}\overline{w_{n+1}},
\end{equation}
}}
where $\overline{w}$ denotes the complex conjugate of $w$.
Then via the projective geometry~\cite{goldman1999complex}, the formula of the unit ball model is:\footnote{Here we denote the dimension as $n$ instead of $N$ as in Section \ref{sec:hyperbolic geometry} since the dimensions of the two models can differ in Fourier transform, which we will see in Section \ref{sec: Fourier transform}.} \\
\scalebox{0.90}{\parbox{1.111\linewidth}{
\begin{equation}
\label{eq:unit ball}
    \mathcal{B}_\mathbb{C}^n
    =\{(z_1,\cdots,z_n,1)||z_1|^2+\cdots+|z_n|^2<1\}.
\end{equation}
}}
The metric on $\mathcal{B}_\mathbb{C}^n$ is Bergman metric, which takes the formula below in 2-d case: \\
\scalebox{0.90}{\parbox{1.111\linewidth}{
\begin{equation}
    ds^2=\frac{-4}{\llangle\mathbf{z},\mathbf{z}\rrangle^2}\det\begin{bmatrix}
\llangle\mathbf{z},\mathbf{z}\rrangle & \llangle d\mathbf{z},\mathbf{z}\rrangle\\
\llangle\mathbf{z},d\mathbf{z}\rrangle & \llangle d\mathbf{z},d\mathbf{z}\rrangle
\end{bmatrix}.
\end{equation}
}}
Then the distance function on $\mathcal{B}_\mathbb{C}^n$ can be derived from the metric tensor: \\
\scalebox{0.90}{\parbox{1.111\linewidth}{
\begin{equation}
\label{eq:unit ball distance}
    d_\mathcal{B}(\mathbf{z},\mathbf{w})=\text{arcosh}\Big(2\frac{\llangle\mathbf{z},\mathbf{w}\rrangle\llangle\mathbf{w},\mathbf{z}\rrangle}{\llangle\mathbf{z},\mathbf{z}\rrangle\llangle\mathbf{w},\mathbf{w}\rrangle}-1\Big).
\end{equation}
}}

Although the unit ball model and the Poinca\'e ball model look similar in mathematical {formulae}, they have many differences in properties since the complex hyperbolic geometry and hyperbolic geometry are intrinsically different geometries. Not only the variable/constant negative curvature but also their distance functions and the geometric computations vary with each other.


\begin{figure*}[t]
\begin{center}
{\includegraphics[width=1.5\columnwidth]{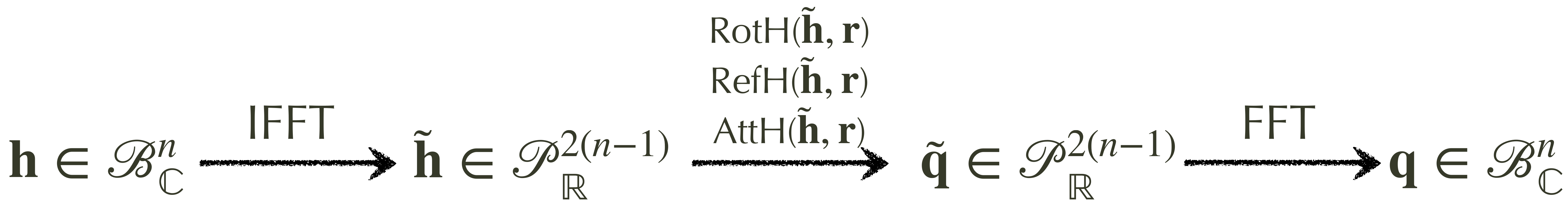}}
\caption{The inference process for a query $(h,r)$ in our proposed complex hyperbolic KG embedding framework. $\mathbf{h}$ and $\tilde{\mathbf{h}}$ are the head entity embeddings in different spaces. $\mathbf{q}$ and $\tilde{\mathbf{q}}$ are the query embeddings in different spaces. $\mathcal{B}_\mathbb{C}^n$ denotes the $n$-dimension unit ball model in the complex hyperbolic space while $\mathcal{P}_\mathbb{R}^{2(n-1)}$ denotes the $2(n-1)$-dimension Poincar\'e ball model in the hyperbolic space.}
\label{fig:inference process}
\end{center}
\end{figure*}

\section{Approach}
Given the KG with entity set $V=\{e_j\}_{j=1}^m$, relation set $R=\{r_j\}_{j=1}^k$, and triplet set $F=\{(h,r,t)|h,t\in V,r\in\ R\}$, the link prediction task aims to predict the tail entity $t$ for each test query $(h,r)$. To train the models for a higher-quality inference, we learn the entity embeddings $\{\mathbf{e}_j\}_{j=1}^m$ in the unit ball model, while parameterizing the relations by the hyperbolic transformations RotH, RefH, and the hyperbolic attention-based model AttH~\cite{DBLP:conf/acl/ChamiWJSRR20}. We construct the conversion between the two geometries through Fourier transform. Figure \ref{fig:inference process} presents the overview of our framework. In this section, we first present the attention-based hyperbolic transformations as relation parameterization, then introduce the Fourier transform as the conversion between complex hyperbolic domain and hyperbolic domain, followed by the details of our framework.

\subsection{Relation Parameterization by Hyperbolic Transformations and Attention}
In our work, we adopt the attention-based hyperbolic transformations developed by~\citet{DBLP:conf/acl/ChamiWJSRR20} as the relation parameterization in the Poincar\'e ball model. Here we present the models RotH, RefH, and AttH.

RotH and RefH represent rotations and reflections in the hyperbolic space respectively. They can be modeled using the Givens transformation matrices, which take the following formula: \\
\scalebox{0.90}{\parbox{1.111\linewidth}{
\begin{equation}
    G^{\pm}(\theta)=\begin{bmatrix}
\cos(\theta) & \mp\sin(\theta) \\
\sin(\theta) & \pm\cos(\theta)
\end{bmatrix}.
\end{equation}
}}
Let $\Theta_r\equiv(\theta_{r,j})_{j\in\{1,\dots,\frac{N}{2}\}}$ and $\Phi_r\equiv(\phi_{r,j})_{j\in\{1,\dots,\frac{N}{2}\}}$ be the relation parameters, where $N$ denotes the dimension of the embedding space, then the hyperbolic rotation and reflection are parameterized by the block-diagonal Givens matrices: \\
\scalebox{0.90}{\parbox{1.111\linewidth}{
\begin{align}
    \label{eq:Rot diagonal transformation}
    &\text{Rot}(\Theta_r)=\text{diag}(G^+(\theta_{r,1}),\dots,G^+(\theta_{r,\frac{N}{2}})), \\
    \label{eq:Ref diagonal transformation}
    &\text{Ref}(\Phi_r)=\text{diag}(G^-(\phi_{r,1}),\dots,G^-(\phi_{r,\frac{N}{2}})).
\end{align}
}}
Thus, given a query $(h,r)$, RotH and RefH apply hyperbolic rotation and reflection with relation-specific parameters $\mathbf{r}$ to the head embeddings $\tilde{\mathbf{h}}\in\mathcal{P}_\mathbb{R}^N$, and then get the query embeddings: \\
\scalebox{0.88}{\parbox{1.136\linewidth}{
\begin{equation}
\label{eq:RotH,RefH}
\begin{small}
    \text{RotH}(\tilde{\mathbf{h}},\mathbf{r})=\text{Rot}(\Theta_r)\tilde{\mathbf{h}}, 
    \text{RefH}(\tilde{\mathbf{h}},\mathbf{r})=\text{Ref}(\Phi_r)\tilde{\mathbf{h}}.
\end{small}
\end{equation}
}}

In order to handle multiple relation properties, AttH combines the above two representations using the hyperbolic attention and adding a hyperbolic translation $\mathbf{r}_r$ by M\"{o}bius addition (Eq. (\ref{eq:Mobius addition})): \\
\scalebox{0.88}{\parbox{1.136\linewidth}{
\begin{equation}
\label{eq:AttH}
\begin{small}
    \text{AttH}(\tilde{\mathbf{h}},\mathbf{r})=\text{Att}(\text{RotH}(\tilde{\mathbf{h}},\mathbf{r}), \text{RefH}(\tilde{\mathbf{h}},\mathbf{r});\mathbf{a}_r)\oplus_{c_r}\mathbf{r}_r,
\end{small}
\end{equation}
}}
where $c_r$ is the curvature parameter of $r$. RotH, RefH, and AttH leverage the trainable curvature so that each relation has its own curvature parameterization. The hyperbolic attention is constructed from the exponential map (Eq. (\ref{eq:exp map})) of the average in the tangent space~\cite{DBLP:conf/nips/ChamiYRL19,DBLP:conf/nips/LiuNK19}. More details about the hyperbolic attention mechanism can be referred to~\cite{DBLP:conf/acl/ChamiWJSRR20}.

\subsection{Conversion by Fourier Transform}
\label{sec: Fourier transform}
The orthonormal Discrete Fourier Transform (DFT) $\mathcal{F}$ and its inverse (IDFT) $\mathcal{F}^{-1}$ between two finite {complex}-valued sequences $\{x_p\}_{p=0}^{N-1}$ and $\{z_q\}_{q=0}^{N-1}$ take the following formulae: \\ 
\scalebox{0.90}{\parbox{1.111\linewidth}{
\begin{align}
\label{eq:DFT}
&z_q=\mathcal{F}\{\mathbf{x}\}_q=\frac{1}{\sqrt{N}}\sum_{p=0}^{N-1}x_p\cdot e^{-i\frac{2\pi}{N}pq}, \\
\label{eq:IDFT}
&x_p=\mathcal{F}^{-1}\{\mathbf{z}\}_p=\frac{1}{\sqrt{N}}\sum_{q=0}^{N-1}z_q\cdot e^{i\frac{2\pi}{N}pq}.
\end{align}
}}

In our models, we transform the unit ball embeddings $\mathbf{z}\in\mathcal{B}_\mathbb{C}^n$ to the Poincar\'e ball embeddings $\mathbf{x}\in\mathcal{P}_
\mathbb{R}^N$ back and forth. Note that the Poincar\'e ball embeddings $\mathbf{x}=\{x_0,\dots,x_{N-1}\}$ are all real numbers, then $\mathcal{F}\{\mathbf{x}\}$ is symmetric: $z_{q}=\overline{z_{-q\text{ mod }N}}, \forall q\in\{0,\dots,N-1\}$. The dimension $N$ is even because of the construction of diagonal Givens transformations (Eqs. (\ref{eq:Rot diagonal transformation}) and (\ref{eq:Ref diagonal transformation})). Then it follows that $z_0$ and $z_{\frac{N}{2}}$ are real-valued, and the remainder of $\mathcal{F}\{\mathbf{x}\}$ is completely specified by just $\frac{N}{2}-1$ complex numbers. Therefore, in practical algorithms, we set $N=2(n-1)$, i.e., we use the first $\frac{N}{2}+1$ elements $\{z_0,\dots,z_{\frac{N}{2}}\}$ as the transformed unit ball embeddings.


We notice that the Fourier transform is not simply a conversion technique between complex and real domains.
{Performing circular convolutions} in one domain {equals} the multiplication in another domain~\cite{DBLP:journals/tc/Rader72, smith1997scientist}: \\
\scalebox{0.90}{\parbox{1.111\linewidth}{
\begin{align}
\label{eq:circular}
\{x \star y\}[n] &\triangleq \sum_{p=0}^{N-1}x_p \cdot y_{(n-p) \text{ mod } N} \nonumber \\
&= \mathcal{F}^{-1}\{\mathcal{F}\{\mathbf{x}\} \cdot \mathcal{F}\{\mathbf{y}\}\}_{n}.
\end{align}
}}
Its effectiveness provides useful transforms, while its practicability is guaranteed by FFT and IFFT, where the fast Fourier algorithms can reduce the computing complexity from $O(N^2)$ to $O(N\log N)$~\cite{cooley1965algorithm}.

\subsection{Complex Hyperbolic Embeddings with Fourier Transform}

For a query $(h,r)$, Figure \ref{fig:inference process} briefly describes the inference process in our framework. The head embedding $\mathbf{h}\in\mathcal{B}_\mathbb{C}^n$ is in the $n$-dimension unit ball model. We apply inverse Fourier transform (Eq. (\ref{eq:IDFT})) to $\mathbf{h}$ and get the transformed head embeddings in the $2(n-1)$-dimension Poincar\'e ball model $\tilde{\mathbf{h}}\in\mathcal{P}_\mathbb{R}^{2(n-1)}$:
$\tilde{\mathbf{h}}=\mathcal{F}^{-1}\{\mathbf{h}\}$.

Then we can apply RotH, RefH (Eq. (\ref{eq:RotH,RefH})), and AttH (Eq. (\ref{eq:AttH})) to get the query embedding $\tilde{\mathbf{q}}$: \\
\scalebox{0.90}{\parbox{1.111\linewidth}{
\begin{equation}
    \tilde{\mathbf{q}}_{\text{ModelH}}=\text{ModelH}(\tilde{\mathbf{h}},\mathbf{r})
\end{equation}
}}
where $\text{ModelH}=\{\text{RotH},\text{RefH},\text{AttH}\}$ represents the corresponding hyperbolic embedding model.

The query embedding $\tilde{\mathbf{q}}$ then gets transformed back to the unit ball model by Fourier transform (Eq. (\ref{eq:DFT})):
$\mathbf{q}=\mathcal{F}\{\tilde{\mathbf{q}}\}$.

Finally, we use the following score function~\cite{DBLP:conf/nips/BalazevicAH19} to measure the likelihood of a triplet $(h,r,t)$: \\
\scalebox{0.90}{\parbox{1.111\linewidth}{
\begin{equation}
    s(h,r,t)=-d_\mathcal{B}(\mathbf{q},\mathbf{t})^2+b_h+b_t,
\end{equation}
}}
where $d_\mathcal{B}(\mathbf{q},\mathbf{t})$ is the unit ball model distance (Eq. (\ref{eq:unit ball distance})) between the tail embeddings $\mathbf{t}$ and the query embeddings $\mathbf{q}$ computed by the above procedures. $b_h$ and $b_t$ are bias terms of the head and tail entity. The model learns the embeddings by maximizing the score functions of the training triplets, i.e., making the query embeddings of $(h,r)$ close with its ground truth tail embeddings. The score function is also used to predict the test data.

In summary, our model parameters include the entity parameters: $\{\mathbf{e}_j\}_{j=1}^m\in\mathcal{B}_\mathbb{C}^n$ (embeddings), $\{b_j\}_{j=1}^m$ (biases); and the relation parameters: 
$\Theta_r$ (rotations), $\Phi_r$ (reflections), $\mathbf{r}_r$ (translations), $\mathbf{a}_r$ (attention), $c_r$ (curvature). The FFT and IFFT can be computed very efficiently, so our models have almost the same computation cost with the base models RotH, RefH, and AttH, while we utilize a more powerful representation geometry to improve the embedding quality.

\section{Experiments}
In this section, we evaluate our approaches on the KG link prediction task.
We show that our complex hyperbolic embedding models outperform the baseline methods based on other geometric spaces.

\subsection{Experimental Settings}
\subsubsection{Data}

\begin{table}[t]
    \centering
    \begin{small}
    \begin{tabular}{l|rrrr}
    \toprule
          & $|V|$ & $|R|$ & $|F|$ &  $\xi_G$     \\
          \midrule
WN18RR    & 40,943         & 11           & 93,003        & -2.54 \\
FB15k-237 & 14,541         & 237          & 310,079       & -0.65 \\
\bottomrule
    \end{tabular}
    \end{small}
    \caption{Data statistics. $|V|$, $|R|$, $|F|$ denote \# entities, \# relations, \# triplets. $\xi_G$ is the global graph curvature.}
    \label{tab:data statistics}
\end{table}

We use two widely-used KG benchmarks to evaluate the embedding models. The data statistics are provided in Table \ref{tab:data statistics}. The global graph curvature $\xi_G$~\cite{DBLP:conf/iclr/GuSGR19} is provided in~\cite{DBLP:conf/acl/ChamiWJSRR20}, which is a distance-based measure to estimate the tree-likeness of graphs.
{A lower $\xi_G$ corresponds to a more tree-like graph.}

\paragraph{WN18RR.} WN18RR~\cite{DBLP:conf/nips/BordesUGWY13} is a knowledge graph dataset created from WN18, which is a subset of WordNet~\cite{DBLP:journals/cacm/Miller95}. WordNet is a large lexical database with hypernymy relation, so WN18RR inherits the underlying hierarchical structure.

\paragraph{FB15k-237.} FB15k-237~\cite{toutanova2015observed} is a knowledge graph dataset created from FB15k, which is derived from Freebase. Compared with WN18RR, FB15k-237 has much more relations and various relation properties, resulting in a more flexible structure, which can be reflected by the larger $\xi_G$.

We follow the train-valid-test data splitting of previous works~\cite{DBLP:conf/acl/ChamiWJSRR20,DBLP:conf/emnlp/WangLLS21}, where \# train-valid-test triplets are $86,845$-$3,034$-$3,134$ for WN18RR and $272,115$-$17,535$-$20,466$ for FB15k-237. The data can be obtained in the public repository of~\cite{DBLP:conf/acl/ChamiWJSRR20}.\footnote{\url{https://github.com/HazyResearch/KGEmb}.}

\subsubsection{Baselines}
The following KG embedding baselines are compared with our approaches (\textbf{FFTRefH}, \textbf{FFTRotH}, \textbf{FFTAttH}): 
complex Euclidean embedding models \textbf{ComplEx-N3}~\cite{DBLP:conf/icml/LacroixUO18} and \textbf{RotatE}~\cite{DBLP:conf/iclr/SunDNT19};
hyperbolic embedding models \textbf{MuRP}~\cite{DBLP:conf/nips/BalazevicAH19}, \textbf{RefH}, \textbf{RotH}, \textbf{AttH}~\cite{DBLP:conf/acl/ChamiWJSRR20}; 
the Euclidean analogues of the hyperbolic methods \textbf{MuRE}, \textbf{RefE}, \textbf{RotE}, \textbf{AttE};
the lightweight Euclidean-based models \textbf{RotL}, \textbf{Rot2L}~\cite{DBLP:conf/emnlp/WangLLS21}.

\subsubsection{Training and Evaluation}
For the baselines, we either take the results from the original papers~\cite{DBLP:conf/acl/ChamiWJSRR20,DBLP:conf/emnlp/WangLLS21} (Table \ref{tab:overall results}) or use their released best hyperparameters as well as their open-source codes to train their models (Table \ref{tab:WN18RR relations}, \ref{tab:WN18RR dimension}, and \ref{tab:FB15K dimension}). For our approaches, we tune the hyperparameters by grid search on each validation set in {$32$-dimension} complex hyperbolic space, which are given in Appendix \ref{app:hyperparameters}. Our embedding models are trained by optimizing the full cross-entropy loss with uniform negative sampling. We conduct all the experiments on four NVIDIA GTX 1080Ti GPUs with {11GB} memory each.

We use the mean reciprocal rank (MRR) and the proportion of correct types that rank no larger than N (Hits@N) as our evaluation metrics, which are widely used for evaluating link prediction. We follow the filtered evaluation setting~\cite{DBLP:conf/nips/BordesUGWY13} to filter out the true triplets during evaluation. In all experiments, each running is executed {five} times and the mean values of results are reported.

\subsection{Overall Results}
\label{sec:overall results}
\begin{table*}[t]
    \centering
    \begin{small}
    \begin{tabular}{l|l|cccc|cccc}
    \toprule
       & & \multicolumn{4}{c|}{WN18RR}     & \multicolumn{4}{c}{FB15k-237}  \\
        \midrule
Geometry & Model   & MRR   & Hits@1 & Hits@3 & Hits@10 & MRR   & Hits@1 & Hits@3 & Hits@10 \\
\midrule
\multirow{2}{*}{$\mathbb{C}^n$} & ComplEx-N3 & 0.420 & 0.390 & 0.420 & 0.460 & 0.294 & 0.211 & 0.322 & 0.463 \\
 & RotatE & 0.387 & 0.330 & 0.417 & 0.491 & 0.290 & 0.208 & 0.316 & 0.458 \\
 \midrule
 \multirow{6}{*}{$\mathbb{R}^n$} & MuRE & 0.458 & 0.421 & 0.471 & 0.525 & 0.313 & 0.226 & 0.340 & 0.489 \\
  & RefE & 0.455 & 0.419 & 0.470 & 0.521 & 0.302 & 0.216 & 0.330 & 0.474 \\
  & RotE & 0.463 & 0.426 & 0.477 & 0.529 & 0.307 & 0.220 & 0.337 & 0.482 \\
  & AttE & 0.456 & 0.419 & 0.471 & 0.526 & 0.311 & 0.223 & 0.339 & 0.488 \\
  & RotL & 0.469 & 0.426 & - & 0.550 & 0.320 & 0.229 & - & 0.500 \\
  & Rot2L & 0.475 & \underline{0.434} & - & 0.554 & \underline{0.326} & \underline{0.237} & - & 0.503 \\
 \midrule
 \multirow{4}{*}{$\mathcal{P}^{n}_\mathbb{R}$} & MuRP & 0.465 & 0.420 & 0.484 & 0.544 & 0.323 & 0.235 & 0.353 & 0.501 \\
 & RefH    & 0.447 & 0.408 & 0.464 & 0.518  & 0.312 & 0.224 & 0.342 & 0.489  \\
 & RotH    & 0.472 & 0.428 & 0.490 & 0.553  & 0.314 & 0.223 & 0.346 & 0.497  \\
 & AttH    & 0.466 & 0.419 & 0.484 & 0.551  & 0.324 & {0.236} & 0.354 & 0.501  \\
\midrule
 \multirow{3}{*}{$\mathcal{B}^n_\mathbb{C}$} & FFTRefH & 0.463 & 0.412 & 0.480 & 0.547  & {0.325} & 0.234 & \underline{0.359} & \underline{0.508}  \\
 & FFTRotH & \textbf{0.484} & \textbf{0.437} & \textbf{0.502} & \textbf{0.572}  & 0.319 & 0.228 & 0.352 & 0.500  \\
 & FFTAttH & \underline{0.476} & {0.432} & \underline{0.494} & \underline{0.558}  & \textbf{0.331} & \textbf{0.239} & \textbf{0.365} & \textbf{0.517} \\
\bottomrule
    \end{tabular}
    \end{small}
    \caption{Evaluation of link prediction task in $32$-dimension embedding spaces. The best results are shown in boldface. The second best results are underlined.}
    \label{tab:overall results}
\end{table*}

\begin{table*}[t]
\centering
\begin{small}
\begin{tabular}{lrrr|rrr|rrr}
\toprule
        Relation & $Khs_G$ & $\xi_G$ & \# Triplets & RefH & RotH & AttH & FFTRefH & FFTRotH & FFTAttH \\
        \midrule
        member meronym & 1.00 & -2.90 & 253 & 0.316 & 0.383 & 0.383 & 0.366 & \textbf{0.411} & 0.402 \\
        hypernym & 1.00 & -2.46 & 1,251 & 0.218 & 0.268 & 0.257 & 0.249 & \textbf{0.283} & 0.268 \\
         has part & 1.00 & -1.43 & 172 & 0.259 & 0.303 & 0.294 & 0.287 & \textbf{0.347} & 0.335 \\
         instance hypernym & 1.00 & -0.82 & 122 & 0.471 & 0.480 & 0.471 & 0.496 & \textbf{0.503} & 0.499 \\
         member of domain region & 1.00 & -0.78 & 26 & 0.417 & 0.417 & 0.404 & \textbf{0.436} & 0.423 & 0.410 \\
        member of domain usage & 1.00 & -0.74 & 24 & 0.424 & 0.451 & 0.445 & 0.431 & \textbf{0.458} & 0.424 \\
        synset domain topic of & 0.99 & -0.69 & 114 & 0.352 & 0.417 & 0.406 & 0.436 & \textbf{0.475} & 0.444 \\
        derivationally related form & 0.07 & -3.84 & 1,074 & 0.960 & 0.964 & 0.965 & 0.968 & \textbf{0.969} & 0.967 \\
        also see & 0.36 & -2.09 & 56 & 0.664 & 0.640 & 0.649 & \textbf{0.684} & 0.675 & 0.676 \\
        similar to & 0.07 & -1.00 & 3 & \textbf{1.000} & \textbf{1.000} & 0.944 & \textbf{1.000} & \textbf{1.000} & \textbf{1.000} \\
        verb group & 0.07 & -0.50  & 39 & \textbf{0.974} & \textbf{0.974} & 0.970 & \textbf{0.974} & \textbf{0.974} & 0.970 \\
\bottomrule
\end{tabular}
\end{small}
 \caption{Results of Hits@10 for WN18RR relations in $32$-dimension embedding spaces. Higher $Khs_G$ and lower $\xi_G$ correspond to more tree-like. \# Triplets means the triplet count of each relation in test set. The best results are shown in boldface.}
    \label{tab:WN18RR relations}
\end{table*}

Table \ref{tab:overall results} presents the results in $32$-dimension embedding spaces. We strictly follow the experimental setting and data splitting of the previous works~\cite{DBLP:conf/acl/ChamiWJSRR20,DBLP:conf/emnlp/WangLLS21}. The results of the baselines are taken from the original papers, where RotL and Rot2L do not report the Hits@3 scores, thus we leave them blank. 

The results show that our Fourier transform-based complex hyperbolic approaches have the best performance on the link prediction task, demonstrating the powerful representation capacity of the complex hyperbolic geometry and the effectiveness of Fourier transform.
Specifically, FFTRotH achieves the best results on WN18RR, while FFTAttH outperforms other methods on FB15k-237. The relations in WN18RR typically have transitivity property, in which case the hyperbolic rotation takes more advantages. FB15k-237 is a more challenging link prediction dataset since it has more relations and varying structures as well as a larger scale of triplets. Therefore, the attention mechanism helps to generalize the hyperbolic transformations to multiple relation properties.

From Table \ref{tab:overall results}, we see that the traditional complex Euclidean models (ComplEx-N3 and RotatE) do not have competitive performance with the hyperbolic KG embedding models or their Euclidean analogues. The hyperbolic methods (MuRP, RefH, RotH, and AttH) have better results than their Euclidean analogues (MuRE, RefE, RotE, and AttE), revealing the improvements of the hyperbolic geometry over Euclidean geometry in low-dimensional KG representation. RotL replaced the M\"{o}bius addition of RotH with a new flexible addition operation, while Rot2L further utilizes two stacked rotation-translation layers in the Euclidean space. The two Euclidean-based methods outperform their base model RotH by adapting a lightweight architecture. However, they still cannot achieve as promising results as the complex hyperbolic embedding approaches.

\subsection{Exploring the Relations}

\begin{table*}[t]
\centering
\begin{small}
\begin{tabular}{l|rr|rr|rr|rr}
\toprule
        & \multicolumn{2}{c|}{$8$-dimension} & \multicolumn{2}{c|}{$16$-dimension} & \multicolumn{2}{c|}{$32$-dimension} & \multicolumn{2}{c}{$64$-dimension} \\
        \midrule
Model   & MRR        & Hits@1      & MRR         & Hits@1      & MRR         & Hits@1      & MRR         & Hits@1      \\
\midrule
RefH    & 0.190      & 0.140      & 0.401       & 0.360      & 0.447       & 0.408      & 0.475       & 0.433      \\
RotH    & 0.220      & 0.154      & 0.417       & 0.370      & 0.472       & 0.428      & \textbf{0.488}       & \textbf{0.442}      \\
AttH    & 0.158      & 0.102      & 0.404       & 0.356      & 0.466       & 0.419      & 0.476       & 0.430      \\
\midrule
FFTRefH & 0.369      & 0.319      & 0.447       & 0.408      & 0.463       & 0.412      & 0.469       & 0.425      \\
FFTRotH & \textbf{0.411}      & \textbf{0.358}      & \textbf{0.468}       & \textbf{0.423}      & \textbf{0.484}       & \textbf{0.437}      & \textbf{0.488}       & \textbf{0.442}      \\
FFTAttH & 0.387      & 0.330      & 0.459       & 0.415      & 0.476       & 0.432      & 0.479       & 0.435     \\
\bottomrule
\end{tabular}
\end{small}
 \caption{Results of MRR and Hits@1 in different embedding dimensions on WN18RR. The best results are shown in boldface.}
    \label{tab:WN18RR dimension}
\end{table*}

\begin{table*}[t]
\centering
\begin{small}
\begin{tabular}{l|rr|rr|rr|rr}
\toprule
        & \multicolumn{2}{c|}{$8$-dimension} & \multicolumn{2}{c|}{$16$-dimension} & \multicolumn{2}{c|}{$32$-dimension} & \multicolumn{2}{c}{$64$-dimension} \\
        \midrule
Model   & MRR        & Hits@1      & MRR         & Hits@1      & MRR         & Hits@1      & MRR         & Hits@1      \\
\midrule
RefH    & 0.267 & 0.188 & 0.288 & 0.204 & 0.312 & 0.224 & 0.328 & 0.237 \\
RotH    & 0.269 & 0.187 & 0.289 & 0.204 & 0.314 & 0.223 & 0.323 & 0.231 \\
AttH    & 0.276 & 0.194 & 0.298 & 0.212 & 0.324 & 0.236 & 0.333 & 0.240  \\
\midrule
FFTRefH & 0.281 & 0.198  & 0.304 & 0.217 & 0.325  & 0.234 & 0.337  & 0.242 \\
FFTRotH & 0.287 & 0.201 & 0.306 & 0.217 & 0.319 & 0.228 & 0.323 & 0.231  \\
FFTAttH & \textbf{0.295} & \textbf{0.209} & \textbf{0.314} & \textbf{0.224} & \textbf{0.331} & \textbf{0.239} & \textbf{0.339} & \textbf{0.245}  \\
\bottomrule
\end{tabular}
\end{small}
 \caption{Results of MRR and Hits@1 in different embedding dimensions on FB15k-237. The best results are shown in boldface.}
    \label{tab:FB15K dimension}
\end{table*}

In Section \ref{sec:overall results}, we see that the overall results of Fourier transform-based complex hyperbolic methods surpass their corresponding hyperbolic methods. Here we explore their {performance} on each relation of WN18RR. For each relation, we give their statistics of Krackhardt hierarchy score ($Khs_G$)~\cite{DBLP:conf/nips/BalazevicAH19} and estimated graph curvature $\xi_G$~\cite{DBLP:conf/nips/ChamiYRL19}. Higher $Khs_G$ and lower $\xi_G$ mean more tree-like, i.e., the relation is more transitive. We report the Hits@10 scores in Table \ref{tab:WN18RR relations}. 

We find that for most relations, FFT complex hyperbolic methods outperform hyperbolic methods significantly. For the transitive relations such as \textit{member meronym}, \textit{hypernym}, \textit{has part}, etc, rotation has much better results than reflection. This phenomenon is consistent with the analysis of previous work~\cite{DBLP:conf/acl/ChamiWJSRR20}, where they found hyperbolic rotations work better on anti-symmetric relations while hyperbolic reflections encode symmetric relations better. Transitivity fulfills anti-symmetry naturally, so rotation gains higher scores (RotH$>$RefH, FFTRotH$>$FFTRefH). For the symmetric relation such as \textit{also see}, reflection outperforms rotation (RefH$>$RotH, FFTRefH$>$FFTRotH). Since most relations in WN18RR exhibit transitivity, the rotation models have better performance than the reflection models in overall results (Table \ref{tab:overall results}). Regardless of the relation properties, our approaches improve the corresponding hyperbolic methods largely, except for the relations with few test triplets such as \textit{similar to} and \textit{verb group}, where they all have close-to-$1$ Hits@10 results.

\subsection{Exploring the Embedding Dimensions}

In this section, we explore the {performance} of Fourier transform-based complex hyperbolic approaches and the corresponding hyperbolic methods in various embedding dimensions. The results are presented in Table \ref{tab:WN18RR dimension} and \ref{tab:FB15K dimension}. We find that when the embedding dimension is small, the complex hyperbolic approaches outperform the hyperbolic base models by a large margin. Remarkably, FFTRotH improves over RotH by around $100\%$ in $8$-dimension on WN18RR. With the increase of the embedding dimension, their {predictions get} more and more similar and gradually converge. The results {reveal} the effectiveness of our approaches especially in small dimensions, demonstrating the strong representation capacity of complex hyperbolic geometry.

\section{Conclusion and Future Work}
In this work, we explore the complex hyperbolic geometry for multi-relational KG embeddings. The whole framework utilizes the Fourier transform as the efficient conversion between geometric spaces.
With the aid of the Fourier transform, the complex hyperbolic embeddings can be transformed into the real domain and be capable of applying real hyperbolic transformations, which enables our approach to take the advantages of both the powerful complex hyperbolic geometry and the attention-based real hyperbolic transformations. Experiments show that the Fourier transform-based complex hyperbolic embedding models can effectively learn the KG embeddings and outperform the baseline models of other spaces in the link prediction task. We believe our proposed approach not only provides a novel and interesting representation learning framework for KGs but also potentially inspires the learning algorithms for more general multi-relational data and contributes to improvements on more downstream tasks.

\section*{Limitations}
\paragraph{Limited improvements in high dimensions.} Although our approaches can significantly outperform the baselines in low-dimensional KG embedding setting, we find that our approaches would get converge and have close results with the hyperbolic base models in sufficiently high dimensions. For example, in Table \ref{tab:WN18RR dimension}, FFTRotH and RotH have the same results in $64$-dimension embedding spaces on WN18RR.

This issue has been observed previously~\cite{DBLP:conf/nips/NickelK17,DBLP:conf/acl/ChamiWJSRR20}, though their comparisons are established between hyperbolic space and Euclidean space. The representation capacity gap between geometric spaces is distinctly revealed in low dimensions. The gap may get eliminated to some extent by increasing the dimension. The complex hyperbolic geometry and hyperbolic geometry usually converge their results in much lower dimensions than Euclidean geometry because of the exponential growth property,
resulting in the limited improvements in high dimensions.

\section*{Acknowledgements}
The authors of this paper were supported by the NSFC Fund (U20B2053) from the NSFC of China, the RIF (R6020-19 and R6021-20) and the GRF (16211520) from RGC of Hong Kong, the MHKJFS (MHP/001/19) from ITC of Hong Kong and the National Key R\&D Program of China (2019YFE0198200) with special thanks to HKMAAC and CUSBLT, and  the Jiangsu Province Science and Technology Collaboration Fund (BZ2021065). We also thank the support from the UGC Research Matching Grants (RMGS20EG01-D, RMGS20CR11, RMGS20CR12, RMGS20EG19, RMGS20EG21). We also thank the support from NVIDIA AI Technology Center (NVAITC).


\bibliography{Reference}
\bibliographystyle{acl_natbib}

\newpage
\appendix

\section{Hyperparameters}
\label{app:hyperparameters}
\begin{table*}[t]
    \centering
    \begin{small}
    \begin{tabular}{l|l|lrrrr}
        \toprule
        Data & Model & Optimizer & Batch size & Negative samples & Learning rate & Double negative \\
        \midrule
        \multirow{3}{*}{WN18RR} & FFTRefH & Adam & 500 & 100 & 0.0003 & True \\
        & FFTRotH & Adam & 500 & 100 & 0.0003 & True \\
        & FFTAttH & Adam & 500 & 100 & 0.0004 & True \\
        \midrule
        \multirow{3}{*}{FB15k-237} & FFTRefH & Adagrad & 500 & 250 & 0.02 & False \\
        & FFTRotH & Adam & 100 & 100 & 0.0002 & False \\
        & FFTAttH & Adagrad & 500 & 100 & 0.03 & False \\
        \bottomrule
    \end{tabular}
    \end{small}
    \caption{Hyperparameters of our approaches.}
    \label{tab:hyperparameters}
\end{table*}

We tune our hyperparameters by grid search on each validation set in {$32$-dimension} complex hyperbolic space, which are given in Table \ref{tab:hyperparameters}. For FFT and IFFT algorithms, we use the package torch.fft\footnote{\url{https://pytorch.org/docs/stable/fft.html}.} and set the parameter norm=``ortho'', which is consistent with the defined orthonormal Fourier transform in Section \ref{sec: Fourier transform}.

\end{document}